\documentclass[10pt,twocolumn,letterpaper]{article}

\usepackage{wacv}              

\definecolor{wacvblue}{rgb}{0.21,0.49,0.74}
\usepackage{booktabs}    
\usepackage{multirow}    
\usepackage[pagebackref,breaklinks,colorlinks,allcolors=wacvblue]{hyperref}
\usepackage{tikz}
\usepackage{pgfplots}
\usepackage[accsupp]{axessibility}
\pgfplotsset{compat=1.15} 
\usepgfplotslibrary{groupplots}
\usepackage{pifont}
\newcommand{\cmark}{\ding{51}}
\newcommand{\xmark}{\ding{55}}

\def\wacvPaperID{3644} 
\def\confName{WACV}
\def\confYear{2027}

\title{A Two-Stage Shape- and Motion-Aware Framework \\ for mmWave-based Human Mesh Recovery}

\author{
Hoang Hai Pham \quad Shuntian Zheng \quad Jiaqi Li \quad
Yu Guan\thanks{Corresponding author.}\\
University of Warwick, United Kingdom \\
{\tt\small \{hoang-hai.pham, yu.guan\}@warwick.ac.uk}
}

\begin{document}
\maketitle
\begin{abstract}
Millimeter-wave (mmWave) radar has emerged as a promising sensing modality for human perception, offering robustness under challenging environmental conditions and strong privacy preservation. However, recovering accurate 3D human meshes from radar is difficult for two reasons: radar tensors contain significant environmental clutter, and each frame observes only part of the body surface. These challenges demand opposite operations, removing measured energy that does not belong to the body versus inferring geometry the sensor never measured. Yet previous works typically entangle both within a single network, with the mesh objective as the main supervision. We instead reformulate radar-based mesh recovery as two explicitly supervised sub-problems. A signal-level separation stage isolates body-related voxels through coarse-to-fine localization and voxel-wise segmentation, trained
with visibility-derived voxel labels. A structure-level reconstruction stage then recovers the body from this partial observation using a dual-branch network that jointly models per-frame shape and inter-frame motion. Experiments show that our method reduces mesh error by up to 9.1\% over the strongest baseline while using over $10\times$ fewer parameters. Code will be available \href{https://github.com/haipham03/mmTSM-Mesh}{here}.
\end{abstract}
    
\section{Introduction}
\label{sec:intro}

Human mesh recovery (HMR), the task of reconstructing the full 3D human
body from sensor observations, is an important problem with applications
in virtual reality~\cite{virtual_1, virtual_2}, human--robot
interaction~\cite{human_robot_1, human_robot_2}, and health
monitoring~\cite{health_1, health_2}. Unlike human pose estimation, which recovers only sparse joints, HMR
captures the dense body surface and shape, enabling more detailed
modeling. While RGB-based HMR has achieved
substantial progress, vision sensors remain unreliable in dark, smoky, or
privacy-sensitive environments. Millimeter-wave (mmWave) radar has
therefore emerged as a promising alternative, offering robustness under
low illumination, resilience to occlusion, and inherent privacy
preservation through its reliance on radio reflections rather than visual
appearance~\cite{lu2020see, fan2026mmpred, pppr}.

Recovering the human body from mmWave radar, however, faces two key
challenges. The first is \emph{signal-level clutter}. Point-cloud-based
approaches~\cite{mmbat,mmmesh,p4trans} rely on threshold-based detectors
such as CFAR~\cite{finn1968adaptive}, which thresholds signal intensity
without semantic awareness of the underlying source. Weak but
informative body reflections are thus discarded, and the resulting
sparse point clouds lack the fine-grained structure required for dense
reconstruction. Tensor-based approaches~\cite{m4human, millimamba,
rtpose} preserve these weak reflections by operating on the raw radar
tensor. However, most voxels correspond to environmental clutter,
multipath artifacts, or empty space. These methods must suppress
clutter implicitly within a single end-to-end network, relying
on weak, indirect supervision from the final reconstruction
objective. The second challenge is \emph{partial observation}. Radar
measures only the surfaces facing the antenna. Each frame therefore
observes a subset of the body, and self-occluded regions produce no
measurable return. Existing approaches typically resort to simple
feature aggregation across frames. They lack mechanisms that recover
unobserved geometry from shape priors or motion cues, which are
important for human-centric tasks~\cite{choi2025mvdoppler, milliflow}.

Our key insight is that these two challenges demand opposite
inference operations. Clutter suppression requires discarding
information, removing measured energy that does not belong to the
body, whereas partial observation requires adding information,
inferring geometry from priors because the sensor never measured it.
A framework supervised only by the final mesh loss must perform
both at once, with one indirect signal that never indicates which
operation failed. We therefore reformulate radar HMR as two
explicitly supervised sub-problems: signal-level separation and
structure-level reconstruction. The first stage isolates
body-related voxels via coarse-to-fine localization and
voxel-wise segmentation, trained with visibility-derived labels marking body surface visible from the radar, and outputs a body-likelihood volume. The second stage reconstructs the full mesh from
this incomplete evidence with a dual-branch network, in which a
shape branch encodes per-frame geometry under the guidance of a
complete-body shape prior, while a motion branch captures body dynamics across the input sequence. The
fused shape and motion features are decoded into SMPL-X parameters. In summary, our contributions are as follows:
\begin{itemize}
    \item We reformulate mmWave-based HMR as two explicitly
    supervised sub-problems, signal-level separation and
    structure-level reconstruction, replacing the weak, indirect
    supervision of a single mesh loss.

    \item For separation, we introduce a coarse-to-fine reflection
extraction module trained with visibility-derived voxel labels,
which also improves existing methods as a drop-in preprocessor.

    \item For reconstruction, we design a dual-branch network with a
shape branch guided by a complete-body prior and a motion branch
capturing inter-frame dynamics, enabling robust mesh recovery under
partial observation.

    \item Experiments under cross-subject and cross-action protocols show consistent gains over existing methods, with up to 9.1\% lower mesh error at over 10$\times$ fewer parameters, and a cross-dataset evaluation shows lower joint error under environment shift.
\end{itemize}
\section{Related Work}

\subsection{mmWave-based Human Perception}

MmWave-based methods commonly adopt one of two input representations: point clouds and radar tensors. Point clouds are obtained by applying CFAR~\cite{finn1968adaptive} to radar tensors, in which reflections are selected via threshold-based detection. While many CFAR variants~\cite{os-cfar,cfar_sea,cfar-ca} have been proposed to improve this process, such heuristic pipelines are not tailored for downstream human reconstruction and inevitably discard fine-grained signals related to body structure. Recent works have therefore explored using the radar tensor directly as input. RT-Pose~\cite{rtpose} introduces HRRadarPose, a single-stage architecture that extracts high-resolution representations from the raw radar tensor in 3D space for pose estimation. RETR~\cite{retr} extends the DETR framework~\cite{detr} to multi-view radar heatmaps for indoor object detection and instance segmentation, introducing a depth-prioritized positional encoding and a tri-plane bounding-box loss spanning both radar views and the image plane. These methods demonstrate that radar tensors retain richer information than CFAR-extracted point clouds, but at the cost of severe environmental clutter and multipath artifacts that the network must handle.

Beyond the spatial structure encoded in point clouds and radar tensors, motion cues have proven valuable for radar-based human sensing. Kim and Ling~\cite{svm_doppler} show that micro-Doppler signatures alone are sufficient to classify human activities with over 90\% accuracy, while MVDoppler-Pose~\cite{choi2025mvdoppler} more recently exploits radial velocity for pose estimation under self-occlusion. However, Doppler is limited to 1D motion along the line of sight and is unavailable in certain radar configurations such as the Vayyar sensor~\cite{vayyarhome}. As a Doppler-free alternative, MilliFlow~\cite{milliflow} estimates 3D scene flow directly from radar point clouds and demonstrates that this information improves downstream performance. These advances show that motion cues are an important feature for precise human-centric tasks.

\subsection{mmWave-based Human Mesh Recovery}
A human body mesh is a triangulated surface with thousands of vertices,
making direct vertex regression impractical. Existing works therefore
adopt parametric body models such as SMPL-X~\cite{SMPL-X}, which control
the full mesh through low-dimensional pose and shape parameters, reducing
mesh recovery to parameter regression.

For point clouds as the input, mmMesh~\cite{mmmesh} introduces the first real-time mmWave-based human mesh estimation system, distributing anchor points around the body to extract local features from sparse point clouds for SMPL-X parameter regression. P4Transformer~\cite{p4trans} is a generic point cloud video architecture commonly adopted as a backbone for spatiotemporal radar modeling. mmBaT~\cite{mmbat} predicts subject translations to crop body points for each frame, then regresses SMPL parameters from the cropped points via a skeleton-aware coarse-to-fine estimator. While effective, these methods all rely on CFAR-filtered point clouds as input and thus inherit the information loss of threshold-based detection. mmDEAR~\cite{mmdear} instead attempts to mitigate this sparsity by densifying the input point cloud through a multi-stage completion network supervised by 2D human masks from synchronized RGB images; however, this requires paired image data during training, which is costly to obtain
and limits applicability to image-free deployments. More recently, Fan~\etal~\cite{m4human} introduce M4Human, to date
the first and only benchmark pairing raw radar tensors with mesh
annotations, and propose the first method for mesh recovery directly from radar tensors, demonstrating that operating on the raw tensor preserves richer information for fine-grained reconstruction. However, existing methods largely treat clutter suppression and body reconstruction within a single end-to-end pipeline under only the
indirect supervision of the mesh objective, while also overlooking
shape priors and motion cues in temporal data.
\begin{figure*}[!t]
    \centering
    \includegraphics[width=0.87\textwidth]{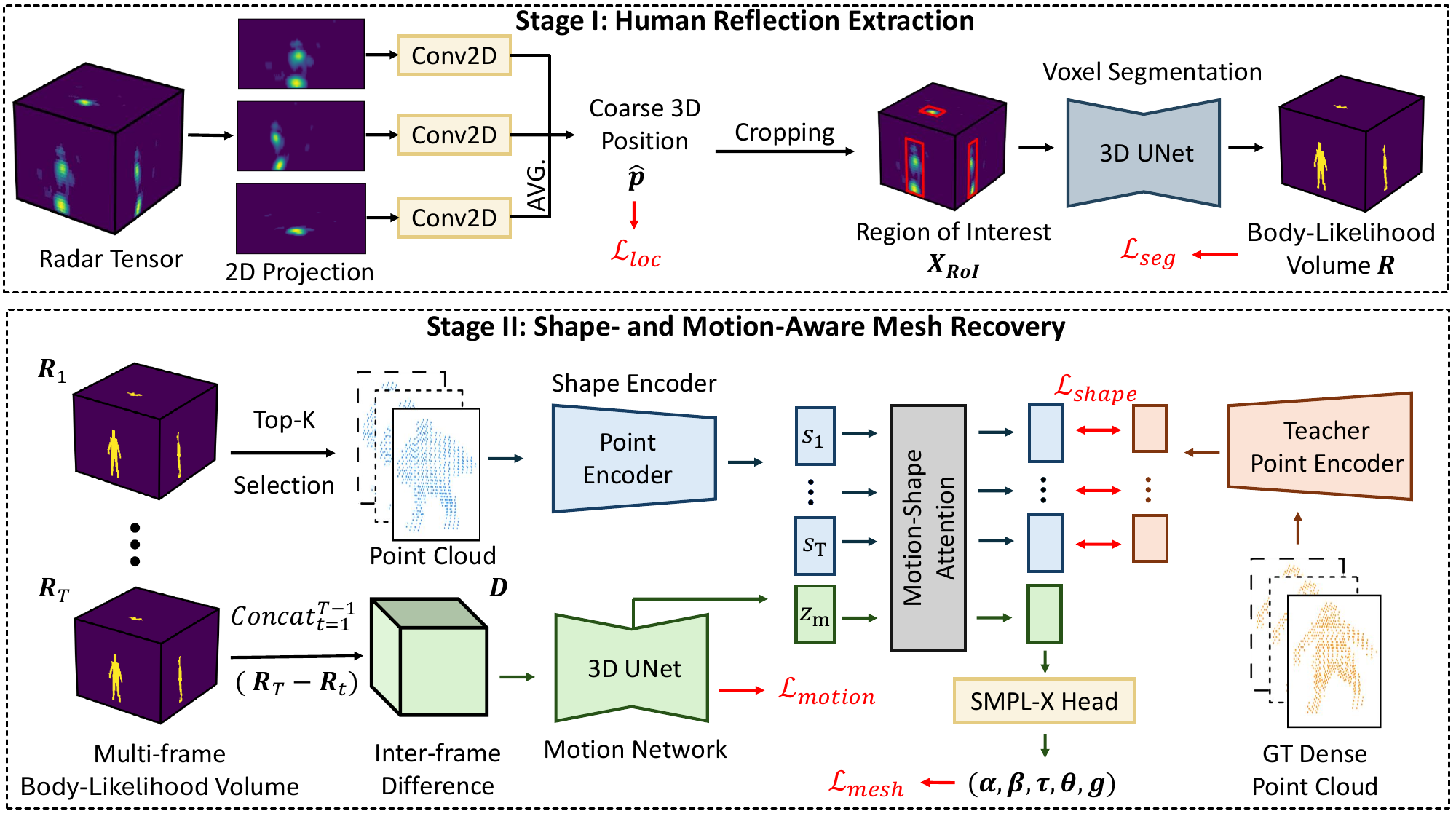}
    \caption{The proposed framework consists of a Human Reflection Extraction (HRE) stage and a Shape- and Motion-Aware Mesh Recovery (SMMR) stage. HRE is first trained independently and then used as a preprocessing step for both training and inference of the SMMR.}
    \label{fig:overview}
\end{figure*}

\section{Methods}
 
\subsection{Problem Formulation}

After standard signal processing, an mmWave radar frame is commonly represented as a tensor over Range, Azimuth, Elevation, and Doppler, $(R, A, E, D)$~\cite{rtpose, hupr}. However, the Doppler dimension is unavailable on some radar
systems, notably high-resolution imaging
radars~\cite{vayyarhome, milliflow, m4human} whose fine spatial
detail suits dense mesh recovery. To make our formulation independent of the specific sensor configuration, we therefore use the spatial $(R, A, E)$ representation shared across radar systems and resample it into a Cartesian grid. Given the resulting temporal sequence $\mathbf{X}_{\text{RT}} \in \mathbb{R}^{T \times X \times Y \times Z}$, where $T$ is the number of frames and $(X, Y, Z)$ are the spatial dimensions, our goal is to regress the SMPL-X parameters $(\alpha, \beta, \tau, \theta, g)$ of the latest frame. Here, $\alpha$ denotes the root orientation in axis-angle representation, $\beta$ the body shape, $\tau$ the global translation, $\theta$ the body pose, and $g$ the gender probability used to select the appropriate male or female SMPL-X template.

\subsection{Overview}
Figure~\ref{fig:overview} provides an overview of the pipeline. Following the reformulation in Section~\ref{sec:intro}, the two stages address the challenges through separate supervision: the first suppresses clutter energy, while the second infers geometry that is not directly observed by the sensor. The Human Reflection Extraction (HRE) stage performs signal-level separation, producing a body-likelihood volume from the input tensor. Here, \emph{reflection} refers to voxels on the radar-visible body surface, which serve as a geometric proxy for measured mmWave returns. The Shape- and Motion-Aware Mesh Recovery (SMMR) stage performs structure-level reconstruction, modeling per-frame shape and inter-frame motion through a dual-branch design to regress the SMPL-X parameters.
 
\subsection{Human Reflection Extraction}
 
The radar tensor spans a large spatial region, while the human body occupies only a small subset of it. To reduce unnecessary computation, we adopt a coarse-to-fine strategy. A lightweight module first estimates a coarse 3D position of the human in the scene, which is then used to crop a fixed-size Region of Interest (RoI). Within this RoI, a voxel-wise segmentation network predicts a per-voxel probability map, indicating the likelihood that each voxel lies on the radar-visible body surface. This design restricts the processing to the relevant region and suppresses distant clutter.
 
\paragraph{Coarse localization.}
Given a single-frame radar tensor, we
extract three orthogonal 2D projections along the X--Y, Y--Z, and
X--Z planes to estimate a coarse human centroid. We use these 2D
projections instead of a full 3D network because they have lower
computational complexity while still being sufficient to estimate
the human position.
 
Each projection is processed by a lightweight 2D CNN that predicts the 2D human centroid in that view, yielding $(\hat{x}_{XY}, \hat{y}_{XY})$, $(\hat{y}_{YZ}, \hat{z}_{YZ})$, and $(\hat{x}_{XZ}, \hat{z}_{XZ})$. Since each spatial axis appears in two projections, we average the two predictions per axis to obtain the coarse 3D position $\hat{\mathbf{p}} = (\hat{x}, \hat{y}, \hat{z}) \in \mathbb{R}^3$:
\begin{equation}
\hat{x} = \tfrac{\hat{x}_{XY} + \hat{x}_{XZ}}{2}, \quad
\hat{y} = \tfrac{\hat{y}_{XY} + \hat{y}_{YZ}}{2}, \quad
\hat{z} = \tfrac{\hat{z}_{YZ} + \hat{z}_{XZ}}{2}.
\end{equation}
We then crop a fixed-size RoI $\mathbf{X}_{\text{RoI}} \in \mathbb{R}^{X' \times Y' \times Z'}$ around $\hat{\mathbf{p}}$. Here, the RoI size is chosen to be sufficiently large to fully contain a person, such that minor localization errors do not impact downstream processing.
 
\paragraph{Voxel-wise segmentation.}
Given the RoI, the goal is to identify the voxels on the body surface visible from the radar. We formulate this as a binary segmentation problem, where $1$ denotes a radar-visible body-surface voxel and $0$ denotes empty space, clutter, or occluded body regions. Inspired by the success of UNet-like architectures~\cite{unet, unet++, valanarasu2022unext} for segmentation in the image domain, we adopt a lightweight 3D variant for this task. The segmentation network $f$ produces a per-voxel probability map
$\mathbf{R} = f(\mathbf{X}_{\mathrm{RoI}}) \in [0,1]^{X' \times Y' \times Z'}$,
indicating the likelihood that each voxel belongs to the radar-visible
body surface.
 
\paragraph{Ground-truth construction.}
Supervision for the voxel-wise segmentation is derived from the ground-truth SMPL-X mesh. Since mmWave radar only observes body surfaces visible from the radar viewpoint, we first identify the subset of mesh vertices visible from that viewpoint. Specifically, given mesh vertices $\mathbf{V} \in \mathbb{R}^{N \times 3}$, we apply Hidden Point Removal~\cite{katz2007direct}, a visibility-based algorithm that removes points occluded from a given viewpoint, to retain only the visible vertices $\mathbf{V}_{\text{vis}}$. The visible vertices are then voxelized within the radar tensor grid to obtain a binary occupancy map $\mathbf{G} \in \{0,1\}^{X \times Y \times Z}$, where each voxel is labeled as $1$ if it contains at least one visible vertex and $0$ otherwise. Supervising against visible surfaces rather than the full mesh approximates the radar observation process, in which occluded body parts produce no direct return.

\paragraph{Training objective.}
The training objective combines a coarse-stage localization loss with a fine-stage segmentation loss. The coarse stage is supervised by the squared L2 distance between the predicted centroid $\hat{\mathbf{p}}$ and the ground-truth centroid $\mathbf{p}_{\text{gt}}$:
\begin{equation}
\mathcal{L}_{\text{loc}} = \| \hat{\mathbf{p}} - \mathbf{p}_{\text{gt}} \|_2^2.
\end{equation}
The fine-stage segmentation is supervised by a combination of Binary Cross-Entropy and Dice losses applied within the RoI:
\begin{equation}
\mathcal{L}_{\text{seg}} = \mathcal{L}_{\text{BCE}}({\mathbf{R}}, \mathbf{G}_{\text{RoI}}) + \mathcal{L}_{\text{Dice}}({\mathbf{R}}, \mathbf{G}_{\text{RoI}}),
\end{equation}
where $\mathbf{R}$ is the predicted voxel-wise probability map and
$\mathbf{G}_{\text{RoI}}$ denotes the ground-truth occupancy map cropped
to the RoI. The Binary Cross-Entropy term provides per-voxel supervision, while the Dice term encourages overlap with the ground-truth mask. The total loss for Stage 1 is:
\begin{equation}
\mathcal{L}_{\text{HRE}} = \mathcal{L}_{\text{loc}} + \lambda_{\text{seg}} \mathcal{L}_{\text{seg}}.
\end{equation}
 
\subsection{Shape- and Motion-Aware Mesh Recovery}
The preceding stage suppresses environmental clutter and extracts
body-related voxels, producing a body-likelihood volume that highlights regions likely to correspond to the human body. However, this volume remains fundamentally incomplete: only a subset of the body surface is observed in any given frame, with self-occluded regions producing no measurable return. Moreover, the per-frame output captures no motion cues, which prior work has shown to be valuable for human-centric perception~\cite{choi2025mvdoppler, milliflow, doppler}. To address these challenges, we propose a dual-branch network consisting of a shape branch that encodes per-frame spatial geometry and a motion branch that captures inter-frame body dynamics. The motion branch is supervised by ground-truth scene flow, while the shape branch is guided by a teacher network trained on complete body surface points. The features from both branches are fused through a motion-shape attention module before regressing the SMPL-X parameters.
 
\paragraph{Shape branch.}
For each frame $t \in \{1, \ldots, T\}$, we select the top-$K$ voxels with the highest likelihood from the body-likelihood volume $\mathbf{R}_t$ and convert them to 3D points, forming a sparse point cloud $\mathbf{P}_t$. Each point carries $5$ features: its 3D position, the radar intensity, and the likelihood score. Each frame $\mathbf{P}_t$ is independently processed by a shared point encoder and globally pooled to obtain a per-frame shape token $\mathbf{s}_t$, producing a sequence of shape tokens $[\mathbf{s}_1, \mathbf{s}_2, \ldots, \mathbf{s}_T]$. These tokens capture the spatial geometry of the body as observed in each frame, but lack explicit information about how the body evolves across time.
 
\paragraph{Motion branch.}
To capture temporal dynamics, we model motion as changes in the body-likelihood volume across the $T$ input frames. Taking the latest frame $T$ as the reference, and re-cropping all $\mathbf{R}_t$ with the RoI of frame $T$ so that voxels correspond across frames (Supp.~\ref{sec:supp_imp}), we compute its differences against all preceding frames:
\begin{equation}
\mathbf{D} = \mathrm{Concat}_{t=1}^{T-1} \left( \mathbf{R}_T - \mathbf{R}_t \right).
\end{equation}
By aggregating differences across multiple frames, $\mathbf{D}$ provides rich temporal context for motion estimation. Static regions tend to cancel out in the difference, while moving body parts produce strong responses, making motion cues more explicit.
The resulting difference volume $\mathbf{D}$ is processed by a lightweight 3D UNet-like network. The encoder extracts hierarchical motion features, which are globally pooled to produce a motion token $\mathbf{z}_m$ summarizing the overall body dynamics. In parallel, the decoder predicts a dense scene flow volume $\hat{\mathbf{F}} \in \mathbb{R}^{3 \times X' \times Y' \times Z'}$, representing per-voxel displacement. This prediction is supervised using ground-truth mesh vertex motion, encouraging the network to learn physically meaningful motion representations. At inference time, the decoder branch is discarded, and only the motion token $\mathbf{z}_m$ is retained for downstream fusion.
 
\paragraph{Motion-Shape Attention.}
The motion summary token $\mathbf{z}_m$ is prepended to the shape token sequence, forming the input to a self-attention module:
\begin{equation}
[\mathbf{z}_{\text{fused}}, \mathbf{s}'_1, \mathbf{s}'_2, \ldots, \mathbf{s}'_T] = \operatorname{Attn}\!\left([\mathbf{z}_m, \mathbf{s}_1, \mathbf{s}_2, \ldots, \mathbf{s}_T]\right).
\end{equation}
Each token can attend to all others, producing an output sequence in which each token mixes shape and motion information across the input window. We take $\mathbf{z}_{\text{fused}}$, the output at the motion-token position, as the fused representation, and pass it to the SMPL-X regression head.
 
\paragraph{Training objective.}
To guide both branches toward learning meaningful representations, we introduce two auxiliary losses. First, to supervise the shape branch, we jointly train a teacher network composed of a point encoder and an SMPL-X regression head on dense point clouds sampled from the complete ground-truth meshes. Because the teacher observes the entire body surface, it learns a latent representation of the body's holistic geometry. The post-attention shape tokens $\mathbf{s}'_t$ are then encouraged to align with the corresponding teacher representations through an MSE distillation loss:
\begin{equation}
\mathcal{L}_{\text{shape}} = \frac{1}{T} \sum_{t=1}^{T} \left\| \mathbf{s}'_t - \operatorname{sg}\!\left(\mathbf{s}^{\text{teacher}}_t\right) \right\|_2^2,
\end{equation}
where $\operatorname{sg}(\cdot)$ denotes the stop-gradient operation. Second, to encourage the motion branch to capture meaningful temporal dynamics, it is supervised with a ground-truth scene flow volume via an MSE loss:
\begin{equation}
\mathcal{L}_{\text{motion}} = \| \hat{\mathbf{F}} - \mathbf{F}_{\text{gt}} \|_2^2.
\end{equation}
The ground-truth scene flow volume $\mathbf{F}_{\text{gt}}$ is constructed from per-vertex displacements between the last two ground-truth meshes (frames $t=T-1$ and $t=T$). Each displacement is voxelized into the same grid as the motion input; displacements within the same voxel are averaged, and empty voxels are set to zero. The total Stage 2 objective is:
\begin{equation}
\mathcal{L}_{\text{SMMR}} = \mathcal{L}_{\text{mesh}} + \lambda_{\text{shape}} \mathcal{L}_{\text{shape}} + \lambda_{\text{motion}} \mathcal{L}_{\text{motion}},
\end{equation}
where $\mathcal{L}_{\text{mesh}}$ supervises the predicted SMPL-X
parameters against the ground truth (detailed in
Supp.~\ref{sec:supp_imp}).
\section{Experiments}

\subsection{Experimental Setup} 

\paragraph{Dataset and Protocol.}
We use the M4Human dataset~\cite{m4human} as our primary benchmark, as it is to date the only publicly available dataset for this task. It contains 50 actions, 20 subjects, 999 sequences, and 661K frames collected in indoor environments. The dataset is collected using a Vayyar radar~\cite{vayyarhome}, which does not provide Doppler velocity measurements due to its signal transmission design. We evaluate our method under two protocols: cross-subject and cross-action. In the cross-subject setting, 15 subjects are used for training, 1 for validation, and 4 for testing. In the cross-action setting, 37 actions are used for training, 3 for validation, and 10 for testing.

To evaluate generalization ability, we conduct experiments on the MilliFlow dataset~\cite{milliflow}, which contains 36K frames collected from 12 subjects performing 5 actions across three different scenes: Hallway, Square, and Parking Lot. 
However, since MilliFlow provides only CFAR-processed point clouds, we reconstruct pseudo-radar tensors by projecting the point clouds onto their corresponding locations in the radar volume. We apply the same projection procedure to M4Human to ensure a consistent input representation across both datasets.
The model is trained on M4Human and evaluated on MilliFlow, enabling assessment in unseen environments and domain shifts from indoor to outdoor settings. A detailed discussion of this protocol is in Supp.~\ref{sec:supp_dataset}.

\paragraph{Implementation Details.} We use $T = 4$ consecutive frames as input, following previous work~\cite{m4human}, where each frame has spatial dimensions $(121, 111, 31)$ in the used dataset. The RoI cropped around the predicted human centroid has size $(X', Y', Z') = (32, 32, 24)$. We adopt UNeXt~\cite{valanarasu2022unext} for the voxel segmentation network and motion network, and PointNet++~\cite{qi2017pointnet++} for the point encoder. Our framework is trained in two stages using the Adam optimizer. The first stage is trained for 50 epochs with an initial learning rate of $1 \times 10^{-3}$, decayed by a factor of $0.5$ every 10 epochs. After the first stage, the top $K = 512$ voxels with the highest likelihood are selected to form the point cloud input to the shape branch of the second stage. The second stage is trained for 25 epochs with an initial learning rate of $2 \times 10^{-4}$, decayed by a factor of $0.5$ every 5 epochs. The batch size is $32$ with $L_2$ regularization of $1 \times 10^{-3}$. For the loss weighting, we set $\lambda_{\text{motion}} = 500$, $\lambda_{\text{shape}} = 10$, and $\lambda_{\text{seg}} = 1 \times 10^{-2}$. Additional details are provided in Supp.~\ref{sec:supp_imp}.

\begin{table*}[t]
\centering
\label{tab:comparison}
\resizebox{0.9\textwidth}{!}{
\begin{tabular}{llccccccccccc}
\toprule
\multirow{2}{*}{\textbf{Modality}} & \multirow{2}{*}{\textbf{Method}} & \multicolumn{4}{c}{\textbf{Cross-Subject}} & \multicolumn{4}{c}{\textbf{Cross-Action}} & \multicolumn{3}{c}{\textbf{Efficiency}} \\
\cmidrule(lr){3-6} \cmidrule(lr){7-10} \cmidrule(lr){11-13}
& & \textbf{MVE}$\downarrow$ & \textbf{MJE}$\downarrow$ & \textbf{MRE}$\downarrow$ & \textbf{TE}$\downarrow$ & \textbf{MVE}$\downarrow$ & \textbf{MJE}$\downarrow$ & \textbf{MRE}$\downarrow$ & \textbf{TE}$\downarrow$ & \textbf{Params} & \textbf{GFLOPs} & \textbf{Latency (ms)}\\
\midrule
\multirow{3}{*}{RPC} & mm-Mesh~\cite{mmmesh}  & 170.1 &  147.9 &  16.0 &  100.0 & 173.8 &  146.9 &  16.4 & 91.8 &  41.5M & 2.9 & 4.2\\
                      & P4Trans~\cite{p4trans}  & 140.8 & 125.2 & 15.0 &  86.4 & 147.8 &  126.4 &  16.1 &  78.4 & 129.0M &  11.8 & 10.6\\
                      & mmBaT~\cite{mmbat}  & 159.3 & 138.3 & 16.4 & 91.8 & 169.0 & 143.4  & 17.0  & 88.4  & 37.2M & 2.6 & 6.7\\
\midrule
\multirow{5}{*}{RT}   & RT-Pose~\cite{rtpose}  & 148.1 & 131.2 & 15.3 &  93.9 & 152.8 &   131.0 &  16.0 &  84.6 &  6.1M &  50.7 & 24.6\\
                      & RETR~\cite{retr}     & 169.7 & 148.6 &  16.3 & 100.1 & 163.1 & 138.3 & 16.6 &  85.4 & 52.4M & 3.0 & 11.8 \\
                      & MVDoppler-Pose~\cite{choi2025mvdoppler}     & 146.3 & 126.9 & 15.2  & 94.0 & 148.6 & 130.6 & 16.3 &  77.0 & 8.0M & 9.1 & 12.2\\
                      & milliMamba~\cite{millimamba}     & 151.8 & 134.2  & 15.0 & 97.8 & 156.1 & 134.6 & 16.0 &  84.2 & 10.3M & 47.7 & 22.3\\
                      & RT-Mesh~\cite{m4human}  & 135.1 & 120.2 &  14.9 &  \textbf{86.1} & 143.1 & 122.0 & 15.6 &  76.0 &  63.3M & 2.6 & \textbf{2.7}\\
\midrule
RT                    & Ours & \textbf{128.5} & \textbf{116.0} & \textbf{14.3} & 86.9 & \textbf{130.1} & \textbf{110.1} & \textbf{15.5} & \textbf{73.9} &  \textbf{4.4M} & \textbf{1.7} & 7.5\\
\bottomrule
\end{tabular}
}
\caption{Comparison with existing methods under cross-subject and cross-action protocols on M4Human Dataset~\cite{m4human}.}
\label{tab:comparison}
\end{table*}

\paragraph{Evaluation Metric.} We evaluate reconstruction quality using four commonly adopted metrics, computed directly in the world coordinate frame without root normalization or Procrustes alignment. These operations factor out global translation, rotation, or scale discrepancies between prediction and ground truth, thereby masking errors that are critical for dense human reconstruction. Distances are reported in millimeters, while rotation errors are given in degrees. Specifically, we report Mean Vertex Error (MVE), Mean Joint Error (MJE), Mean Rotation Error (MRE), and Translation Error (TE). 

Note that for the cross-dataset evaluation on MilliFlow, only MJE is reported, as only joint positions are provided in this dataset. We additionally report model parameters (Params), floating-point operations (GFLOPs), and latency to assess computational efficiency. All are measured on the inference pipeline with batch size of $1$, without any caching.

\paragraph{Compared Methods.} 
We benchmark our method against recent radar-based approaches using either radar point clouds (RPC) or radar tensors (RT). All methods are adapted to take multi-frame inputs and use the same number of frames during training and inference. More details about each baseline and the corresponding adaptations are provided in Supp.~\ref{sec:supp_baseline_adaption}.

\begin{table*}[t]
\centering
\resizebox{0.6\textwidth}{!}{
\begin{tabular}{llcccc}
\hline
\textbf{Modality} & \textbf{Method} 
& \textbf{Hallway}
& \textbf{Square}
& \textbf{Parking Lot}
& \textbf{Average} \\
\hline
\multirow{3}{*}{RPC} 
& mm-Mesh~\cite{mmmesh}   & 240.1 & 227.3 & 376.2 & 281.2 \\
& P4Trans~\cite{p4trans}  & 260.0 & 237.1 & 256.5 & 251.2 \\
& mmBaT~\cite{mmbat}      & 249.1 & 234.3 & 253.9 & 245.8 \\
\hline
\multirow{5}{*}{Pseudo-RT} 
& RT-Pose~\cite{rtpose}   & 228.1 & 226.0 & 256.9 & 237.0 \\
& RETR~\cite{retr}        & 230.8 & 226.4 & 251.8 & 236.3 \\
& MVDoppler-Pose~\cite{choi2025mvdoppler}        & 216.2 & 217.8 & 262.2 & 232.1 \\
& milliMamba~\cite{millimamba}        & 215.8 & 222.5 & 242.4 & 226.9 \\
& RT-Mesh~\cite{m4human}  & 214.0 & 207.8 & 231.6 & 217.8 \\
\hline
Pseudo-RT & Ours         & \textbf{207.1} & \textbf{196.8} & \textbf{212.9} & \textbf{205.6} \\
\hline
\end{tabular}
}
\caption{Cross-dataset MJE evaluation on the MilliFlow Dataset~\cite{milliflow}.}
\label{tab:cross_scene}
\end{table*}

\subsection{Quantitative Results}

Table~\ref{tab:comparison} compares our method against eight baselines under
both protocols (single run; std in Supp.~\ref{sec:supp_ablations}). Our method achieves the best MVE, MJE, and MRE on both, and the
best TE on cross-action. Against RT-Mesh, the strongest existing baseline, we
reduce MVE by 9.1\% and MJE by 9.8\% on the more challenging cross-action
protocol, and by 4.9\% and 3.5\% on cross-subject, where TE is comparable. Gains
in MRE are smaller (0.6$^\circ$ and 0.1$^\circ$) but consistent across protocols.

Our method is also efficient: it is over
$10\times$ smaller than RT-Mesh and nearly $30\times$ smaller than P4Trans, with the
lowest compute in the table ($30\times$ below RT-Pose). At 7.5\,ms per sample it
is slower than RT-Mesh (2.7\,ms) but remains comfortably real-time. Per-module breakdowns are in Supp.~\ref{sec:supp_breakdown}.

Table~\ref{tab:cross_scene} reports cross-dataset MJE on MilliFlow. Our method
achieves the lowest MJE in all three scenes (205.6~mm average vs.\ 217.8~mm for
RT-Mesh), with the largest gain in the outdoor Parking Lot scene (18.7~mm).
RPC-based methods degrade most under domain shift, while pseudo-RT methods consistently outperform their RPC counterparts. Since pseudo-RT tensors carry the same CFAR detections as the point clouds, this gap reflects the voxel representation and architectures rather than additional raw-signal information. Our further gain over the pseudo-RT baselines suggests that explicit extraction also reduces sensitivity to scene-specific clutter.

\begin{figure*}[t]
    \centering
    \includegraphics[width=0.7\linewidth]{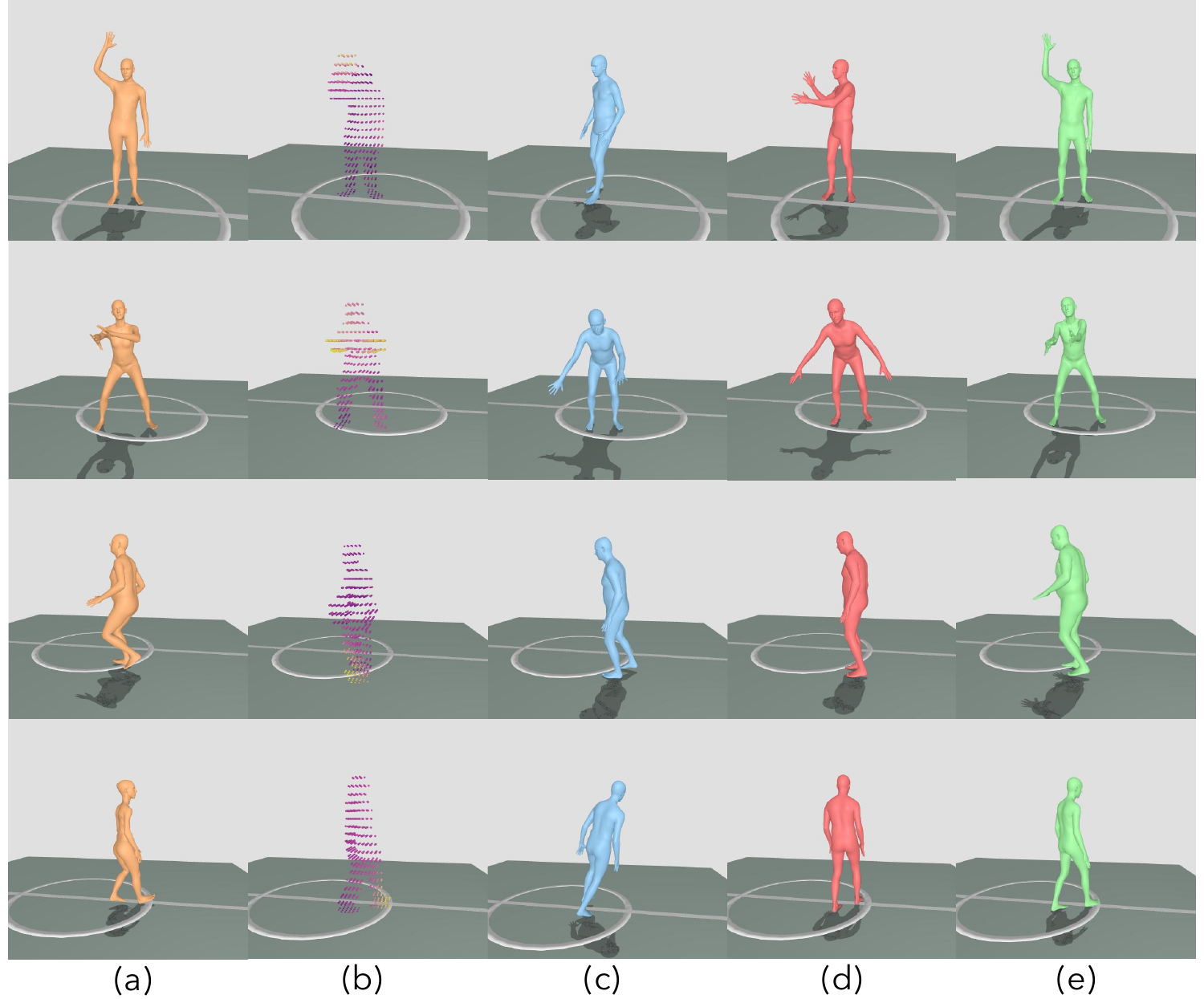}
    \caption{Qualitative comparison of human mesh recovery on the cross-action protocol. 
    (a) Ground-truth mesh; 
    (b) Points extracted by our HRE module, colored by motion magnitude in the motion branch (dark indicates more static points, bright indicates points with larger motion); 
    (c) P4Trans; 
    (d) RT-Mesh; 
    (e) Ours.}
    \label{fig:hmr_qualitative}
\end{figure*}

\subsection{Qualitative Results}
Figure~\ref{fig:hmr_qualitative} presents qualitative comparisons on representative frames from the cross-action protocol, covering a diverse range of poses. The visualizations in column~(b) show that the proposed HRE module effectively suppresses environmental clutter while preserving body-related voxels. In addition, the motion branch produces strong responses around moving limbs, providing informative temporal cues for the downstream mesh regressor.

Comparing the recovered meshes, both P4Trans~(c) and RT-Mesh~(d) show visible
discrepancies from the ground truth, while
our method~(e) aligns more closely across all examples. More visualizations for
the cross-dataset protocol are in
Supp.~\ref{sec:supp_visualization_milliflow}.

\subsection{Ablation Study}

This section analyzes the contributions of the key components in our framework:
the HRE stage and the SMMR stage, together with their robustness under clutter and
self-occlusion. All experiments follow the cross-action protocol unless stated
otherwise. The Supplementary Material provides additional ablations on design
choices (Supp.~\ref{sec:supp_ablations}), a direct evaluation of
HRE extraction quality (Supp.~\ref{sec:supp_hre_direct}), and a
discussion of current limitations and future extensions to multi-person scenarios and
Doppler-enabled systems (Supp.~\ref{sec:supp_future}).

\begin{table}[t]
\centering
\small
\resizebox{0.75\linewidth}{!}{
\begin{tabular}{l|ccc}
\toprule
\textbf{Method} & \textbf{w/o HRE} & \textbf{w/ HRE} & \textbf{Improve} \\
\midrule
P4Trans & 147.8 & 136.6 & 7.6\% \\
RT-Mesh  & 143.1 & 134.2 & 6.2\% \\
Ours & 141.9 & 130.1 & 8.3\%\\
\bottomrule
\end{tabular}
}
\caption{Effect of HRE on different methods.}
\label{tab:stage1_effect}
\end{table}

\paragraph{Effect of Human Reflection Extraction.}
We first evaluate whether the proposed HRE stage benefits not only our method but also existing radar-based mesh recovery approaches. Table~\ref{tab:stage1_effect} compares MVE with and without HRE applied as a preprocessing step for P4Trans, RT-Mesh, and our method. To accommodate the different input formats of each baseline, we adapt HRE accordingly: for P4Trans, we feed the top 512 points with the highest HRE likelihood scores as input; for RT-Mesh, we append the likelihood score as an additional channel alongside the original intensity channel; for our method, the ``w/o HRE'' setting uses CFAR-processed points for the shape branch and the original radar tensor for the motion branch. All three methods improve substantially with HRE, confirming that it is broadly useful as a drop-in preprocessing module. Our method obtains the largest relative gain (8.3\%), suggesting that the cleaner point cloud is best exploited when paired with our second stage. 

\begin{table}[t]
\centering
\small
\resizebox{0.75\linewidth}{!}{
\begin{tabular}{cc|c}
\toprule
\textbf{Motion Branch} & \textbf{Shape Distillation} & \textbf{MVE} \\
\midrule
\xmark & \xmark & 138.0 \\
\cmark & \xmark & 134.2\\
\xmark & \cmark & 133.4 \\
\cmark & \cmark & 130.1 \\
\bottomrule
\end{tabular}
}
\caption{Ablation study on SMMR components.}
\label{tab:ablation}
\end{table}

\paragraph{Effect of Shape- and Motion-Aware Mesh Recovery Components.}
We ablate the two key components of our SMMR stage: the motion branch and the teacher--student shape distillation.
As shown in Table~\ref{tab:ablation},  adding either
component alone improves accuracy by a comparable margin, showing that
inter-frame dynamics provide information beyond per-frame geometry and that the
teacher's complete-body representations inject useful holistic priors despite
the partiality of radar observations. Combining both performs best, indicating
that the two cues play distinct roles: motion supplies temporal dynamics that
distillation cannot, while distillation injects geometric priors that motion
alone cannot recover.

\begin{table}[t]
\centering
\small
\begin{tabular}{lccc}
\hline
Method & Low & Medium & High \\
\hline
Ours          & \textbf{116.4} & \textbf{126.4} & \textbf{150.9} \\
Ours w/o HRE  & 124.9 & 137.4 & 167.8 \\
\hline
\end{tabular}
\caption{MVE (mm) $\downarrow$ under increasing clutter.}
\label{tab:clutter}
\end{table}

\paragraph{Robustness under Clutter.}
To test whether HRE helps most where the motivation predicts, we partition the
test set by signal-to-clutter ratio (SCR),
\begin{equation}
\mathrm{SCR} =
\frac{\frac{1}{|\mathcal{S}|}\sum_{v \in \mathcal{S}} |\mathbf{X}_{\mathrm{RT}}(v)|}
     {\frac{1}{|\mathcal{N}|}\sum_{v \in \mathcal{N}} |\mathbf{X}_{\mathrm{RT}}(v)|},
\end{equation}
where $|\mathbf{X}_{\mathrm{RT}}(v)|$ is the reflection magnitude of the input
radar tensor at voxel $v$, $\mathcal{S}$ is the set of visible body-surface
voxels given by $\mathbf{G}$, and $\mathcal{N}$ the remaining voxels within the
ground-truth RoI. Samples are split into Low (highest 25\% SCR),
Medium (middle 50\%), and High (lowest 25\%) clutter subsets
(Table~\ref{tab:clutter}). Accuracy degrades for both variants as clutter
increases, confirming that residual clutter is a major source of error, but the
benefit of HRE grows with it: MVE improves by 6.8\% on the Low subset, 8.0\% on
Medium, and 10.1\% on High. The spread between the easiest and hardest subsets
also narrows from 42.9~mm without HRE to 34.5~mm with it. When the body
reflection already dominates the RoI, background energy has little effect on
mesh regression; as the body signal weakens, an end-to-end model must separate
signal from clutter implicitly while also regressing shape and pose.
Supervising this separation directly keeps accuracy more stable across clutter
levels.

\begin{table}[t]
\centering
\small
\begin{tabular}{lccc}
\hline
Method & Low & Medium & High \\
\hline
Ours                        & \textbf{109.3} & \textbf{126.8} & \textbf{157.3} \\
Ours w/o motion \& distill. & 114.3 & 134.0 & 169.9 \\
\hline
\end{tabular}
\caption{MVE (mm) under increasing self-occlusion.}
\label{tab:occlusion}
\end{table}

\paragraph{Robustness under Self-occlusion.}
We similarly partition the test set by self-occlusion ratio, computed from the
number of mesh vertices remaining after Hidden Point Removal
(Table~\ref{tab:occlusion}). The gain from the motion and distillation
components grows with occlusion severity, reducing MVE by 4.4\% on the Low
subset, 5.4\% on Medium, and 7.4\% on High, and the degradation from Low to
High is smaller with them (48.0~mm) than without (55.6~mm). This indicates that
the two components recover geometry that is not directly observed in a single
frame, matching their design intent of combining shape priors and inter-frame
motion to compensate for incomplete observations.

\subsection{Failure Analysis}
We trace severe failures (MVE$>$200\,mm) to their source  (details in Supp.~\ref{sec:supp_fail_analysis}). Mesh error is flat for HRE localization errors up to 250\,mm, since the RoI absorbs typical localization noise, and rises monotonically beyond this point. We attribute each severe failure in order: to localization if its error is $\geq$300\,mm, otherwise to segmentation if its Dice is $\leq$0.7, and otherwise to SMMR. Under this order, 45.9\% of severe failures stem from localization, 45.3\% from segmentation, and only 8.8\% arise in SMMR. The latter mostly occur when multiple poses are consistent with the same visible-surface points, an inherent ambiguity of single-view radar observations.

\section{Conclusion}

We presented a two-stage shape- and motion-aware framework for mmWave-based human mesh recovery that decouples radar signal processing from geometric reconstruction. The first stage formulates clutter suppression as a voxel-wise segmentation problem with coarse-to-fine localization, producing a body-likelihood volume from raw radar observations. The second stage addresses the partial-observation challenge of radar by employing a dual-branch network that jointly models per-frame geometry and inter-frame motion, supervised via teacher-student shape distillation and an auxiliary scene flow objective. Experiments demonstrate consistent improvements over existing methods under cross-subject and cross-action settings, and lower joint error in cross-dataset transfer, while using over an order of magnitude fewer parameters and lower computational cost than the strongest baselines. Ablation studies further confirm the effectiveness of the motion and shape components and of explicit visible-surface supervision, whose first stage also serves as a general preprocessing module for existing methods. Limitations and future extensions to multi-person scenarios and Doppler-enabled systems are discussed in Supp.~\ref{sec:supp_future}.


\paragraph{Acknowledgments.}
This work was supported by the SJTU--Warwick Joint PhD Scholarship.

{
    \small
    \bibliographystyle{ieeenat_fullname}
    \bibliography{main}
}

\clearpage

\setcounter{section}{0}
\setcounter{subsection}{0}
\setcounter{figure}{0}
\setcounter{table}{0}
\setcounter{equation}{0}

\renewcommand{\thesection}{\Alph{section}}
\renewcommand{\thesubsection}{\thesection.\arabic{subsection}}
\renewcommand{\thefigure}{S\arabic{figure}}
\renewcommand{\thetable}{S\arabic{table}}
\renewcommand{\theequation}{S\arabic{equation}}

\twocolumn[%
\centering
{\LARGE\bfseries Supplementary Material\par}
\vspace{0.5em}
\vspace{1em}
]

\section{More Implementation Details}
\label{sec:supp_imp}

All experiments are conducted on a NVIDIA RTX 3060 (12\,GB) running Ubuntu 24.04.2 LTS with CUDA 12.1.

\subsection{HRE Implementations}

\paragraph{Coarse localization head.}
The localization head operates on three 2D projections (XY, YZ, ZX) of the
radar tensor, obtained by max-pooling along each spatial axis. Each is
processed by three 2D CNN layers, followed by adaptive average pooling
and a fully-connected head predicting the 2D center in
that plane. The six coordinates are averaged per axis to give
$\hat{\mathbf{p}}$; the target $\mathbf{p}_{gt}$ in
$\mathcal{L}_{\mathrm{loc}}$ is the mask centroid.

\paragraph{Voxel segmentation network.}
A 3D UNeXt~\cite{valanarasu2022unext} on the cropped RoI with base width 8 and 4 encoder stages. A final $1{\times}1{\times}1$ convolution and sigmoid yield the voxel-wise probability map.

\paragraph{Visibility supervision.}
Because the radar observes only front-facing surfaces, Stage~1 is supervised
with visibility masks rather than the full mesh. Ground-truth SMPL-X vertices are transformed into the radar frame and passed
to the HPR operator of Katz~\etal~\cite{katz2007direct}
with the viewpoint at the radar origin, using inversion radius
$R = C\max_i\lVert v_i\rVert$ with $C = 200$. The visible vertices are voxelized into the radar tensor grid to form the binary segmentation target.

\subsection{SMMR Implementations}

\paragraph{Input construction for Stage 2.}
HRE is trained on single frames and applied independently to each of the
$T$ frames at inference. All $T$ probability maps are then cropped with the
RoI of the latest frame (probabilities outside set to zero), so
$\mathbf{R}_1,\dots,\mathbf{R}_T$ share one spatial grid and voxel-wise correspondence holds in $\mathbf{D}$ (Eq.~5 of the main paper). 

\paragraph{Shape branch.}
The point encoder is a PointNet++~\cite{qi2017pointnet++} variant on 512
points per frame: two set-abstraction layers downsample to 64 and 16 points, followed by a per-point
MLP and pooling into a per-frame shape token $\mathbf{s}_t \in \mathbb{R}^{256}$.

\paragraph{Teacher network.}
The teacher shares the point-encoder architecture, adds an SMPL-X head, and
takes 1024 points sampled from the complete ground-truth mesh surface. It is
trained jointly with SMMR but optimized only by its own reconstruction loss.

\paragraph{Motion branch.}
The motion branch is a 3D UNeXt with base width 12. Global average pooling on the bottleneck and linear projection gives the motion token
$\mathbf{z}_m \in \mathbb{R}^{256}$; the decoder outputs the scene flow
prediction $\hat{\mathbf{F}}$. Scene-flow targets $\mathbf{v}_T-\mathbf{v}_{T-1}$ are computed per vertex in radar coordinates and voxelized at the frame-$T$ vertex positions within frame $T$'s RoI. Note that differencing against the latest
frame, as in Eq.~5 of the main paper, is informationally equivalent to
differencing consecutive frames: the two sets are related by an
invertible linear map
($\mathbf{R}_T{-}\mathbf{R}_t = \sum_{k=t}^{T-1}(\mathbf{R}_{k+1}{-}\mathbf{R}_k)$),
which the first convolution can absorb through channel mixing. We adopt
the reference-frame form as each slice then directly encodes the
displacement toward frame $T$, whose mesh is regressed.

\paragraph{Motion-shape attention.}
The attention module takes the motion token and the four per-frame shape tokens as input ($T+1=5$ tokens of dimension 256), adds a learnable positional embedding, and applies a 2-layer Transformer encoder with 4 attention heads, feed-forward dimension 512, and pre-norm layer normalization.

\paragraph{Loss function.} In addition to the shape loss $\mathcal{L}_{\text{shape}}$ and
motion loss $\mathcal{L}_{\text{motion}}$ defined in the main paper, the
predicted mesh is supervised with a mesh loss $\mathcal{L}_{\text{mesh}}$, given by the weighted sum
over SMPL-X components:
\begin{equation}
\begin{aligned}
\mathcal{L}_{\mathrm{mesh}}
={}& \lambda_{\theta}\mathcal{L}_{\mathrm{rot}}(\hat{\theta},\theta)
+ \lambda_{\alpha}\mathcal{L}_{\mathrm{rot}}(\hat{\alpha},\alpha)
+ \lambda_{\beta}\lVert\hat{\beta}-\beta\rVert_2^2 \\
&+ \lambda_{\tau}\lVert\hat{\tau}-\tau\rVert_1
+ \lambda_g\mathrm{BCE}(\hat{g},g).
\end{aligned}
\end{equation}
where $\mathcal{L}_{\mathrm{rot}}$ is
a geodesic rotation loss and $\mathrm{BCE}(\cdot)$ is binary cross-entropy. We
set $\lambda_{\theta} = 15$, $\lambda_{\alpha} = 1$, $\lambda_{\beta} = 0.3$,
$\lambda_{\tau} = 10$, and $\lambda_{g} = 0.5$, following M4Human~\cite{m4human}. Adopting
identical weights ensures that performance differences arise from the model rather than
from loss tuning.

\subsection{Metrics}
We report Mean Vertex Error (MVE), Mean Joint Error (MJE), Mean Rotation Error (MRE), and Translation Error (TE). MVE measures the average point-to-point distance between the predicted mesh and the ground-truth SMPL-X mesh over all 10,475 vertices. MJE calculates the average distance between predicted and reference 3D joints using the 22-joint SMPL-X skeleton. MRE measures
the average differences between predicted joint rotations and the
ground-truth rotations, capturing pose accuracy independently of global position. TE quantifies the distance between the estimated and reference global root translations. Note that for the cross-dataset evaluation on MilliFlow, only MJE is reported, as only joint positions are provided in this dataset. We additionally report model parameters (Params), floating-point operations (GFLOPs), and latency to assess computational efficiency. All are measured on the inference pipeline with batch size of $1$, excluding modules that are used solely during training and discarded at test time.

\section{Adaptation of mmWave-Based Baselines}
\label{sec:supp_baseline_adaption}
We benchmark against recent state-of-the-art methods from both the radar point cloud
and radar tensor modalities.

For point-cloud-based methods, we include mmMesh~\cite{mmmesh}, P4Trans~\cite{p4trans},
and mmBaT~\cite{mmbat}, the three best-performing methods on the mmBody
benchmark~\cite{chen2022mmbody} as reported in mmDEAR~\cite{mmdear}. We exclude mmDEAR
itself, as it requires synchronized RGB images as additional training supervision, which
is unavailable in our setting. All three methods are inherently multi-frame and are used
without modification.

For radar-tensor-based methods, we use RT-Mesh~\cite{m4human} directly, as it was
proposed alongside the M4Human dataset. For the remaining baselines, we make minimal
adaptations that preserve their original modeling intent while aligning them with our
input format. In all cases, we replace the original head with the same
SMPL-X regression head used in our method, so that differences in performance reflect the
backbone rather than the output parameterization.

\paragraph{RT-Pose~\cite{rtpose}.} RT-Pose is a single-frame method that extends HRNet to
3D in order to process the radar cube. To accommodate multi-frame input, we follow the
adaptation strategy of~\cite{m4human} and treat the temporal dimension as the channel
dimension.

\paragraph{RETR~\cite{retr}.} Following~\cite{m4human}, we adapt RETR to extract features
from two complementary views (X-Y BEV and X-Z side view) using a transformer-based
architecture. We retain the original radar tensor feature extractors. Multi-frame input is
handled as in RT-Pose, by treating the temporal dimension as the channel dimension.

\paragraph{MVDoppler-Pose~\cite{choi2025mvdoppler}.} MVDoppler-Pose estimates human pose
from multi-view, multi-modal (spatial and Doppler) input. We retain only the single-view
spatial stream and remove the cross-view and cross-modal fusion modules. Its CNN
and MobileViT modules originally operate on a Range $\times$ Time representation; we
generalize the spatial axis from the 1D range dimension to the full 3D radar volume of our
dataset.

\paragraph{milliMamba~\cite{millimamba}.} milliMamba is a recent multi-frame baseline that
models spatio-temporal dependencies across consecutive radar frames. We remove the
cross-view fusion while retaining the single-view core Mamba encoder, and replace the per-frame
keypoint queries in the STCA decoder with a single mesh query.

\section{Dataset Rationale and Discussion}
\label{sec:supp_dataset}

We adopt M4Human~\cite{m4human} as our primary benchmark, as it is the only public dataset
providing paired radar tensors and ground-truth human meshes. It supports two evaluation
protocols: cross-subject and cross-action. M4Human is, however, captured entirely within a single indoor environment. To additionally
evaluate cross-environment generalization, we adopt MilliFlow~\cite{milliflow}, for two
reasons. First, it spans three distinct environments, allowing us to measure robustness
under environment shift. Second, it uses a radar of the same high resolution as M4Human, so the evaluation is not
additionally confounded by a change in sensor resolution.

The main limitation of MilliFlow is that it releases only
CFAR-processed point clouds rather than raw radar tensors. We
therefore project the point clouds back into the tensor grid to form
a pseudo-RT representation, a strategy adopted as an alternative to
real RT in prior work~\cite{kini2024transhupr, millimamba}. In the
absence of any public cross-environment dataset with real RT, this
is the most rigorous protocol available. Importantly, pseudo-RT is
not simply a sparser version of the raw tensor. CFAR both retains
clutter and discards genuine body returns, so a model must
simultaneously suppress surviving clutter and infer body regions
that produced no detection. To quantify the two error modes, we
compare CFAR occupancy against the ground-truth mesh on M4Human
pseudo-RT: 71.4\% of CFAR-occupied voxels lie more than one voxel
from the visible body surface, while 9.9\% of body-surface voxels
have no detection within one voxel. On MilliFlow, where meshes are
unavailable, we instead measure the bidirectional Chamfer distance
between the CFAR points and the visible ground-truth joints
(Eq.~\ref{eq:cd}): the clutter direction
$\mathcal{P}\!\rightarrow\!\mathcal{J}$ substantially
exceeds the coverage direction $\mathcal{J}\!\rightarrow\!\mathcal{P}$
(Table~\ref{tab:hre_milliflow} first row), consistent with the CFAR visualizations in
Figure~\ref{fig:cross-vis}(b). Residual clutter is therefore
the dominant error mode on both datasets, the same
clutter-suppression challenge that defines the raw-tensor setting.

To directly verify that pseudo-RT is a faithful proxy for comparing methods,
we additionally evaluate the strongest baselines and our method on pseudo-RT
constructed from M4Human, where raw-tensor results are available for
reference. As shown in Table~\ref{tab:pseudo_rt_validation}, the method
ranking is preserved, and all methods degrade slightly. Evaluating on pseudo-RT thus preserves the core
difficulty our method targets, rather than substituting an easier task.

\begin{table}[t]
\centering
\small
\begin{tabular}{lcc}
\hline
Method & Raw RT & Pseudo-RT \\
\hline
MVDoppler-Pose~\cite{choi2025mvdoppler}   & 148.6 & 149.9 \\
RT-Mesh~\cite{m4human}   & 143.1 & 143.7  \\
Ours                     & 130.1 & 130.4 \\
\hline
\end{tabular}
\caption{Validation of the pseudo-RT protocol on M4Human (MVE in mm,
cross-action). The method ranking on pseudo-RT matches that on raw radar
tensors, indicating pseudo-RT is a faithful proxy for relative comparison.}
\label{tab:pseudo_rt_validation}
\end{table}

\section{Visualizations on MilliFlow}
\label{sec:supp_visualization_milliflow}
\begin{figure*}[t]
    \centering
    \includegraphics[width=\linewidth]{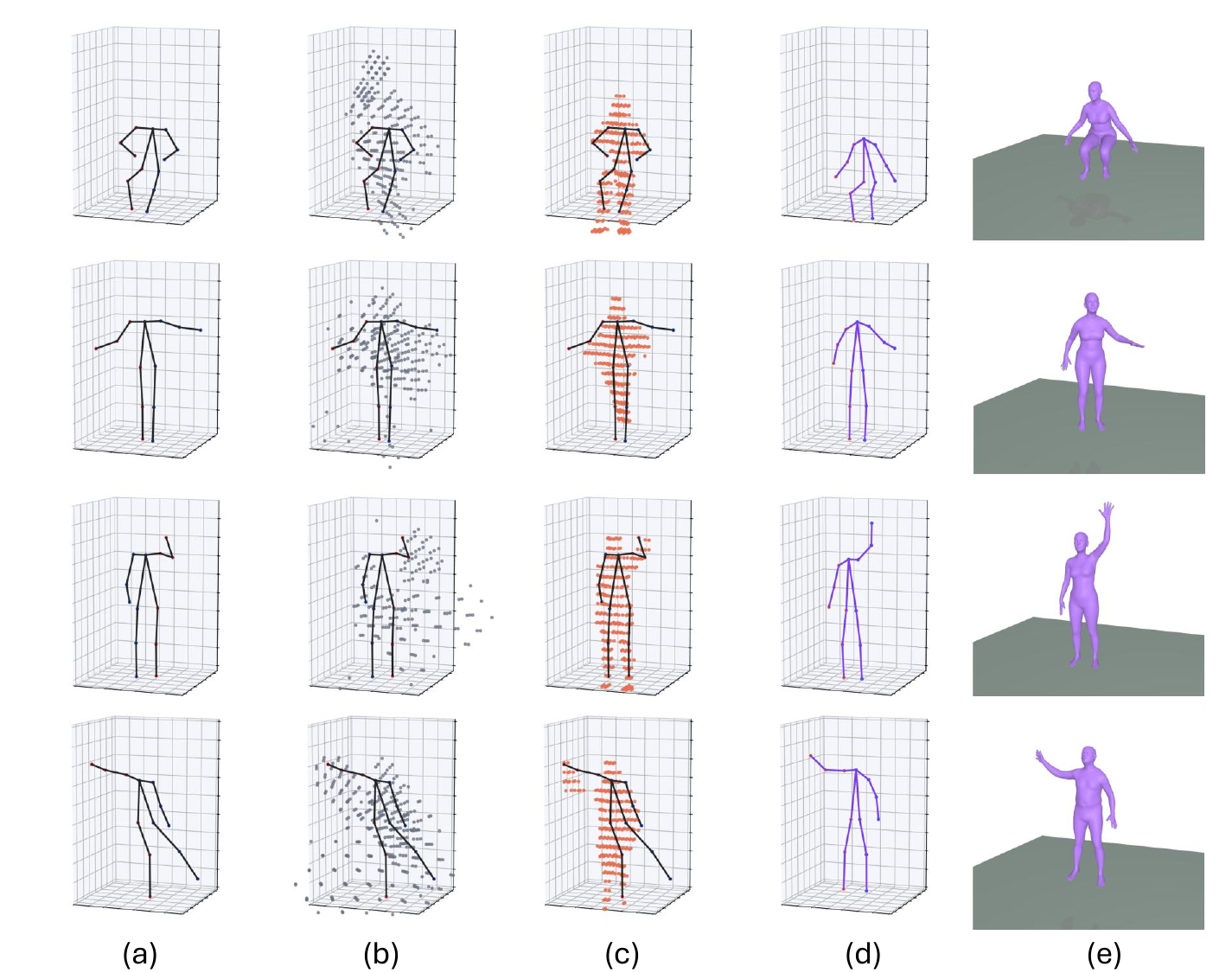}
    \caption{Qualitative results under the cross-dataset protocol on MilliFlow.
    (a) Ground-truth pose. (b) Ground-truth pose with CFAR points. (c) Ground-truth pose
    with HRE points. (d) Predicted pose. (e) Predicted mesh.}
    \label{fig:cross-vis}
\end{figure*}

Figure~\ref{fig:cross-vis} shows example outputs of our method under the cross-dataset
protocol on MilliFlow. The HRE stage discards clutter points away from the body while also
assigning high likelihood to body voxels that CFAR had missed, so the extracted point set follows the body
more closely than the raw CFAR input. Stage~2 is then able to reconstruct a plausible mesh
from these points, despite the pseudo-RT representation and the change of environment.

\section{Efficiency Breakdown}
\label{sec:supp_breakdown}
\begin{table}[t]
\centering
\small
\setlength{\tabcolsep}{5pt}
\renewcommand{\arraystretch}{1.15}
\begin{tabular}{lccc}
\toprule
Module & Params (M) & GFLOPs & Latency (ms) \\
\midrule
\multicolumn{4}{l}{\textbf{Stage 1 (HRE)}}\\
Localization & 0.1 & 0.2 & 0.96 \\
Segmentation & 0.5 & 1.1 & 2.47 \\
\textit{Subtotal} & \textit{0.6} & \textit{1.3} & \textit{3.43} \\
\addlinespace
\multicolumn{4}{l}{\textbf{Stage 2 (SMMR)}}\\
Point Encoder & 0.4 & 0.1 & 1.54 \\
Motion Encoder & 1.2 & 0.3 & 1.42 \\
Self-Attention & 1.3 & $<$0.01 & 0.61 \\
SMPL Head & 0.9 & $<$0.01 & 0.54 \\
\textit{Subtotal} & \textit{3.8} & \textit{0.4} & \textit{4.11} \\
\midrule
\textbf{Total} & \textbf{4.4} & \textbf{1.7} & \textbf{7.54} \\
\bottomrule
\end{tabular}
\caption{Per-module parameter count, compute, and inference latency of our method. Stage 1 rows already account for the T per-frame HRE invocations.}
\label{tab:latency}
\end{table}

Table~\ref{tab:latency} reports the per-module cost of our method. The two stages
have complementary cost profiles, with Stage~1 dominating compute and
Stage~2 holding most of the parameters, while latency is split almost
evenly between them. Coarse localization is inexpensive yet reduces the
segmentation input from the full $121\times111\times31$ tensor to a
$32\times32\times24$ RoI, cutting Stage~1 compute by over $13\times$
(Table~\ref{tab:hre_ablation}).

\section{Extraction Quality of HRE}
\label{sec:supp_hre_direct}
\begin{table}[t]
\centering
\small
\begin{tabular}{lcc}
\toprule
Metric & Cross-subject & Cross-action \\
\midrule
Loc. Err. (mm) $\downarrow$ & 211.6 & 245.7 \\
Precision $\uparrow$        & 83.5 & 82.0 \\
Recall $\uparrow$           & 95.9 & 96.7 \\
Dice $\uparrow$             & 89.3 & 88.7 \\
\bottomrule
\end{tabular}
\caption{HRE performance on M4Human.}
\label{tab:hre}
\end{table}

\begin{table}[t]
\centering
\small
\begin{tabular}{lcccc}
\toprule
Method & Loc. Err. & $\mathcal{P}\rightarrow\mathcal{J}$ & $\mathcal{J}\rightarrow\mathcal{P}$  & $\mathrm{CD}$\\
\midrule
w/o HRE & - & 255.2 & 104.7 & 360.0 \\
HRE  & 178.5 & 162.3 & 61.4 & 223.7\\
\bottomrule
\end{tabular}
\caption{Chamfer Distance (mm) between the extracted point cloud and the visible ground-truth joints on MilliFlow. $\mathcal{P}\rightarrow\mathcal{J}$ and $\mathcal{J}\rightarrow\mathcal{P}$ denote the two directional
components. Lower is better.}
\label{tab:hre_milliflow}
\end{table}

We report the performance of HRE on the M4Human dataset under both protocols, using
localization error, Precision, Recall, and Dice score, as shown in Tab.~\ref{tab:hre}.
HRE performs consistently across protocols, with Dice above 88 and only a small drop from
cross-subject to cross-action. Recall is high in both settings (95.9 and 96.7), so almost
all body-related voxels are retained, while precision is lower (83.5 and 82.0),
indicating that the module keeps some voxels outside the ground-truth surface. This is the
preferable direction for our pipeline: retained non-body voxels are passed to Stage~2 as
additional context and can still be down-weighted there, whereas discarded body voxels
cannot be recovered.

The localization error of 211.6\,mm and 245.7\,mm is well within the RoI crop
($1600\times1600\times2400$\,mm), whose size is set from this error so that the body
remains inside the crop in most cases. The larger error under the cross-action protocol is
consistent with unseen motions producing reflection patterns that shift the estimated body
center, but recall remains high, indicating that Stage~2 still receives essentially
complete input in both settings.

For the cross-dataset protocol, ground-truth meshes are not available, so Precision,
Recall, and Dice cannot be computed. We instead use the Chamfer distance between the
extracted point cloud $\mathcal{P}$ and the set of visible ground-truth joints
$\mathcal{J}$:
\begin{equation}
\label{eq:cd}
\mathrm{CD}(\mathcal{P},\mathcal{J}) =
\underbrace{\frac{1}{|\mathcal{P}|}\sum_{p\in\mathcal{P}}\min_{j\in\mathcal{J}}\lVert p-j\rVert_2}_{\mathcal{P}\rightarrow\mathcal{J}}
+
\underbrace{\frac{1}{|\mathcal{J}|}\sum_{j\in\mathcal{J}}\min_{p\in\mathcal{P}}\lVert j-p\rVert_2}_{\mathcal{J}\rightarrow\mathcal{P}},
\end{equation}
where the first term measures how closely the extracted points follow the body joints and
the second how completely the joints are covered by the extracted points. As shown in
Tab.~\ref{tab:hre_milliflow}, both directions improve, with the larger gain in $\mathcal{P}\rightarrow\mathcal{J}$
(92.9\,mm versus 43.3\,mm), indicating that HRE primarily removes clutter points lying away from the body.

\section{Additional Ablations and Analysis}
\label{sec:supp_ablations}

\subsection{Run-to-Run Variation}
\label{sec:supp_seeds}

To assess training variability, we retrain both stages of our method with $3$ different random seeds under both protocols. Table~\ref{tab:seeds} reports the standard deviation of each metric alongside our improvement over RT-Mesh from Table~1 of the main paper.

\begin{table}[t]
\centering
\resizebox{\columnwidth}{!}{%
\begin{tabular}{l cccc cccc}
\toprule
 & \multicolumn{4}{c}{Cross-Subject} & \multicolumn{4}{c}{Cross-Action} \\
\cmidrule(lr){2-5} \cmidrule(lr){6-9}
 & MVE & MJE & MRE & TE & MVE & MJE & MRE & TE \\
\midrule
Gain vs.\ RT-Mesh & 6.6 & 4.2 & 0.6 & $-0.8$ & 13.0 & 11.9 & 0.1 & 2.1 \\
Std.\ & 0.74 & 0.34 & 0.14 & 0.79 & 0.98 & 0.87 & 0.03 & 0.69 \\
\bottomrule
\end{tabular}}
\caption{Run-to-run standard deviation of our method versus its improvement over RT-Mesh.}
\label{tab:seeds}
\end{table}

\subsection{Effect of Input Sequence Length}

\begin{table}[t]
\small
\centering
\begin{tabular}{lccc}
\hline
Method & $T{=}2$ & $T{=}4$ & $T{=}8$ \\
\hline
P4Trans~\cite{p4trans} & 149.7 & 147.8 & 148.7 \\
RT-Mesh~\cite{m4human} & 145.0 & 143.1 & 143.3 \\
Ours    & 132.8 & 130.1 & 130.4 \\
\hline
\end{tabular}
\caption{Temporal window ablation on cross-action protocol.}
\label{tab:temporal_window}
\end{table}

We ablate the temporal window size, $T \in {2,4,8}$, on the cross-action protocol of the M4Human dataset, with the results reported in Table~\ref{tab:temporal_window}. Across all methods, MVE improves as $T$ increases from 2 to 4, but saturates thereafter, with $T{=}8$ providing no further gain. For our method, increasing the temporal window from $T{=}4$ to $T{=}8$ doubles the computation and latency of the HRE stage compared to the values reported in Table~\ref{tab:latency}, whereas the SMMR stage incurs a negligible overhead of fewer than 0.01M additional parameters and 0.01 GFLOPs.

\subsection{Ablation of Human Reflection Extraction}
To validate the design choices of the Human Reflection Extraction (HRE)
stage, we ablate four factors: the coarse-localization stage, the RoI
size, the segmentation backbone, and the supervision type. For the backbone variants
(UNet~\cite{unet}, UNet++~\cite{unet++}), we adjust the base channel
widths so that the networks have comparable parameter counts. All
variants share the same training schedule, loss, and Stage~2
architecture; only the corresponding Stage~1 component is changed.
Results are reported in Table~\ref{tab:hre_ablation}.

\begin{table}[t]
\small
\centering
\begin{tabular}{lccc}
\toprule
Variant & Params & GFLOPs & MVE\\
\midrule
W/o localization & 0.5M & 17.2 & 132.3 \\
\midrule
RoI $(24,24,16)$ & 0.6M & 0.6 & 134.7 \\
Ours: RoI $(32,32,24)$ & 0.6M & 1.3 & 130.1 \\
RoI $(40,40,32)$ & 0.6M & 2.5 & 129.5 \\
\midrule
UNet~\cite{unet} backbone & 0.7M & 4.5 & 130.7 \\
UNet++~\cite{unet++} backbone & 0.7M & 5.3 & 130.5 \\
Ours (UNeXt~\cite{valanarasu2022unext}) & 0.6M & 1.3 & 130.1 \\
\midrule
Full-mesh supervision & 0.6M & 1.3 & 135.0 \\
Ours (visible-surface) & 0.6M & 1.3 & 130.1 \\
\bottomrule
\end{tabular}
\caption{Ablation of Stage 1 (HRE) design choices on the cross-action
protocol: coarse localization, RoI size, segmentation backbone, and
supervision target. Stage 1 cost is reported separately to highlight the
compute trade-off of each variant.}
\label{tab:hre_ablation}
\end{table}

Removing coarse localization increases MVE by 2.2\,mm and Stage~1 compute
by over $13\times$, since segmentation must process the full
$121{\times}111{\times}31$ tensor rather than a $32{\times}32{\times}24$
RoI; the accuracy drop is likely due to the higher class imbalance at
full-volume scale. Varying the RoI size shows a clear trade-off: smaller crop size increases MVE, as localization errors more frequently
truncate the body, while enlarging it yields marginally lower MVE
 at $2\times$ the compute; we retain
$(32,32,24)$ as the best accuracy--compute balance. Replacing
UNeXt with UNet or UNet++ changes MVE by under 0.5\% but costs  more GFLOPs, so we adopt UNeXt as the cheapest backbone
at comparable accuracy. Finally, supervising with full-mesh occupancy instead of visible-surface labels increases MVE at identical cost. Since occluded voxels produce no return, full-mesh labels require Stage~1 to infer unobserved geometry in addition to separating body voxels from clutter, re-entangling the two operations our decomposition is designed to separate. This ablation supports the benefit of decoupling the two operations under separate supervision.

\begin{table}[t]
\centering
\small
\begin{tabular}{lcccc}
\hline
$\lambda_{\mathrm{seg}}$ & Loc. Err. & Dice \\
\hline
$1\times10^{-3}$          & 243.9 & 88.5\\
$1\times10^{-2}$  & 245.7  & 88.7\\
$1\times10^{-1}$          & 253.6 & 87.9\\
\hline
\end{tabular}
\caption{Sensitivity to the HRE loss weight}
\label{tab:lambda_seg}
\end{table}

\begin{table}[t]
\centering
\small
\begin{tabular}{lccc}
\hline
Variant & MVE \\
\hline
$\lambda_{\mathrm{motion}} = (50/500/5000) $ & (131.2/130.1/131.0)  \\
$\lambda_{\mathrm{shape}} = (1/10/100) $ & (131.8/130.1/130.6) \\
\hline
\end{tabular}
\caption{Sensitivity to the SMMR loss weight}
\label{tab:lambda_smmr}
\end{table}

\subsection{Effect of the Number of Input Points}

\begin{figure}[t]
\centering
\begin{tikzpicture}
\begin{axis}[
    width=5cm,
    height=4.5cm,
    xlabel={Number of Input Points $K$},
    ylabel={MVE (mm)},
    xmin=200, xmax=1080,
    xtick={256, 512, 1024},
    grid=both,
    grid style={line width=.1pt, draw=gray!20},
    tick label style={font=\footnotesize},
    label style={font=\small},
    mark size=3pt,
]
\addplot[color=green!60!black, mark=*, thick] coordinates {
    (256, 134.9)   
    (512, 130.1)   
    (1024, 130.2)  
};
\end{axis}
\end{tikzpicture}
\caption{Effect of the number of selected input points $K$.}
\label{fig:num_points}
\end{figure}

The shape branch selects the top-$K$ voxels as input points. We evaluate $K\in{256,512,1024}$ in Figure~\ref{fig:num_points}. Increasing $K$ from 256 to 512 reduces MVE by 4.8~mm, while further increasing it to 1024 provides no improvement, as lower-confidence points add little useful signal and may introduce clutter. We therefore set $K=512$ for its accuracy and compactness.

\begin{table}[t]
\centering
\small
\begin{tabular}{lc}
\hline
\textbf{Method Variant} & \textbf{MVE} \\
\hline
No teacher & 134.2 \\
Pretrained teacher (frozen) & 131.9 \\
No detach (joint gradient flow) & 133.7 \\
Ours (detached teacher + joint training) & 130.1 \\
\hline
\end{tabular}
\caption{Effect of different training strategies for the shape prior.}
\label{tab:shape_prior_ablation}
\end{table}

\subsection{Effect of the Training Strategy of Shape Prior}
Table~\ref{tab:shape_prior_ablation} compares training strategies for the
teacher--student shape prior: removing the teacher entirely, using a
separately pretrained frozen teacher, joint training with gradients
flowing through both branches, and our design of joint training with
detached teacher embeddings. A frozen pretrained teacher already yields
competitive results, confirming the benefit of the shape prior itself.
Allowing gradients to flow into the teacher degrades performance, likely
because the supervision target becomes unstable when optimized against
the student objective; detaching the teacher embeddings stabilizes the
target while permitting joint training, giving the lowest error.

\subsection{Sensitivity to Loss Weights}
We assess the sensitivity of our framework to the loss weights. For Stage 1,
Table~\ref{tab:lambda_seg} varies $\lambda_{\mathrm{seg}}$ over two orders of
magnitude; both localization error and Dice remain stable across this range.
For Stage 2, Table~\ref{tab:lambda_smmr} varies $\lambda_{\mathrm{motion}}$
and $\lambda_{\mathrm{shape}}$, each while holding the other fixed. MVE
remains stable across all settings, and every variant stays well below the
no-auxiliary-loss baseline (Table~4 of the main paper), so the benefit of the
auxiliary supervision holds across weighting scales.

\subsection{Alignment-Based Evaluation}
\label{sec:supp_pamje}

To analyze the composition of the large cross-dataset errors, we report PA-MJE, which aligns the predicted joints to the
ground truth via Procrustes analysis before computing MJE, removing global
translation, rotation, and scale. As shown in Table~\ref{tab:pamje}, our
method achieves the lowest PA-MJE. For all methods, the aligned error is
well below the unaligned MJE, indicating a sizeable global component in
the cross-dataset error; the residual PA-MJE nonetheless remains above the
in-domain MJE (110.1\,mm), so pose recovery also degrades under
environment shift.

\begin{table}[t]
  \centering
  \begin{tabular}{lcc}
    \hline
    Method & MJE & PA-MJE \\
    \hline
    P4Trans~\cite{p4trans} & 251.2 & 145.3 \\
    RT-Mesh~\cite{m4human} & 217.8 & 130.9 \\
    Ours                   & \textbf{205.6} & \textbf{126.1} \\
    \hline
  \end{tabular}
  \caption{PA-MJE (mm) on MilliFlow, averaged over scenes.}
  \label{tab:pamje}
\end{table}


\begin{figure*}[t!]
\centering
\small
\resizebox{\linewidth}{!}{%
\begin{tikzpicture}
\begin{groupplot}[
    group style={
        group size=4 by 1,
        horizontal sep=2.2cm,
    },
    width=5.5cm,
    height=6cm,
    xlabel={Dataset Scale (\%)},
    xmin=15, xmax=110,
    xtick={25, 50, 100},
    grid=both,
    grid style={line width=.1pt, draw=gray!20},
    legend style={
        at={(0.5,-0.25)},
        anchor=north,
        legend columns=-1,
        font=\large,
        /tikz/every even column/.append style={column sep=1cm},
    },
    tick label style={font=\small},
    label style={font=\large},
    title style={font=\large\bfseries}
]
\nextgroupplot[ylabel={MVE (mm)}, title={MVE vs. Dataset Scale}, legend to name=grouplegend]
\addplot[color=blue, mark=*, thick] coordinates {(25,162.8) (50,156.9) (100,147.8)};
\addlegendentry{P4Trans (RPC)}
\addplot[color=red, mark=square*, thick] coordinates {(25,158.4) (50,149.2) (100,143.1)};
\addlegendentry{RT-Mesh (RT)}
\addplot[color=green!60!black, mark=triangle*, thick, dashed, mark options={solid}] coordinates {(25,144.5) (50,136.7) (100,130.1)};
\addlegendentry{Ours}
\nextgroupplot[ylabel={MJE (mm)}, title={MJE vs. Dataset Scale}]
\addplot[color=blue, mark=*, thick] coordinates {(25,139.8) (50,133.0) (100,126.4)};
\addplot[color=red, mark=square*, thick] coordinates {(25,134.4) (50,125.2) (100,122.0)};
\addplot[color=green!60!black, mark=triangle*, thick, dashed, mark options={solid}] coordinates {(25,123.6) (50,116.2) (100,110.1)};
\nextgroupplot[ylabel={MRE (deg.)}, title={MRE vs. Dataset Scale}]
\addplot[color=blue, mark=*, thick] coordinates {(25,16.7) (50,16.3) (100,16.1)};
\addplot[color=red, mark=square*, thick] coordinates {(25,16.4) (50,15.9) (100,15.6)};
\addplot[color=green!60!black, mark=triangle*, thick, dashed, mark options={solid}] coordinates {(25,16.1) (50,15.8) (100,15.5)};
\nextgroupplot[ylabel={TE (mm)}, title={TE vs. Dataset Scale}]
\addplot[color=blue, mark=*, thick] coordinates {(25,90.5) (50,83.1) (100,78.4)};
\addplot[color=red, mark=square*, thick] coordinates {(25,81.9) (50,77.3) (100,76.0)};
\addplot[color=green!60!black, mark=triangle*, thick, dashed, mark options={solid}] coordinates {(25,86.6) (50,79.5) (100,73.9)};
\end{groupplot}
\node at (group c2r1.south east) [yshift=-1.8cm, xshift=1.1cm] {\textcolor{black}{\ref*{grouplegend}}};
\end{tikzpicture}%
}
\caption{Performance comparison of RT-Mesh, P4Trans, and Ours on the cross-action protocol across different dataset scales.}
\label{fig:dataset_scale_curves}
\end{figure*}

\subsection{Data Efficiency}
\label{sec:supp_data_efficiency}
A key advantage of decomposing the task is that each stage solves a 
simpler sub-problem and therefore requires less data to learn. To 
quantify this, we train our method and the two strongest baselines 
(P4Trans and RT-Mesh) on 25\%, 50\%, and 100\% of the
training set under the cross-action protocol, and report performance 
in Figure~\ref{fig:dataset_scale_curves}. Our method consistently outperforms both baselines across all dataset 
scales and all four metrics. More importantly, the performance gap is largest at 25\% training data: 
with only a quarter of the data, our method achieves 144.5~mm MVE, 
comparable to RT-Mesh's 143.1~mm trained on the full dataset, indicating that our framework reaches the performance of 
state-of-the-art baselines using approximately 25\% of the training 
data they require. This data efficiency is a direct consequence of 
the decomposition: the first stage learns a focused voxel-wise 
classification problem, while the second stage operates on cleaner 
inputs and is further regularized by the shape and motion priors. 
Together, these factors reduce the dependence on large-scale paired 
training data, which is particularly valuable in radar-based human 
sensing where collecting paired radar-mesh data is expensive.

\section{Failure Analysis}
\label{sec:supp_fail_analysis}
Failures of our method originate from three sources: localization error and segmentation
error in the HRE stage, and mesh recovery error in the SMMR stage. We analyze each in turn, with particular attention to how errors propagate from the first stage to the second.

\begin{figure*}
    \centering
    \includegraphics[width=1\linewidth]{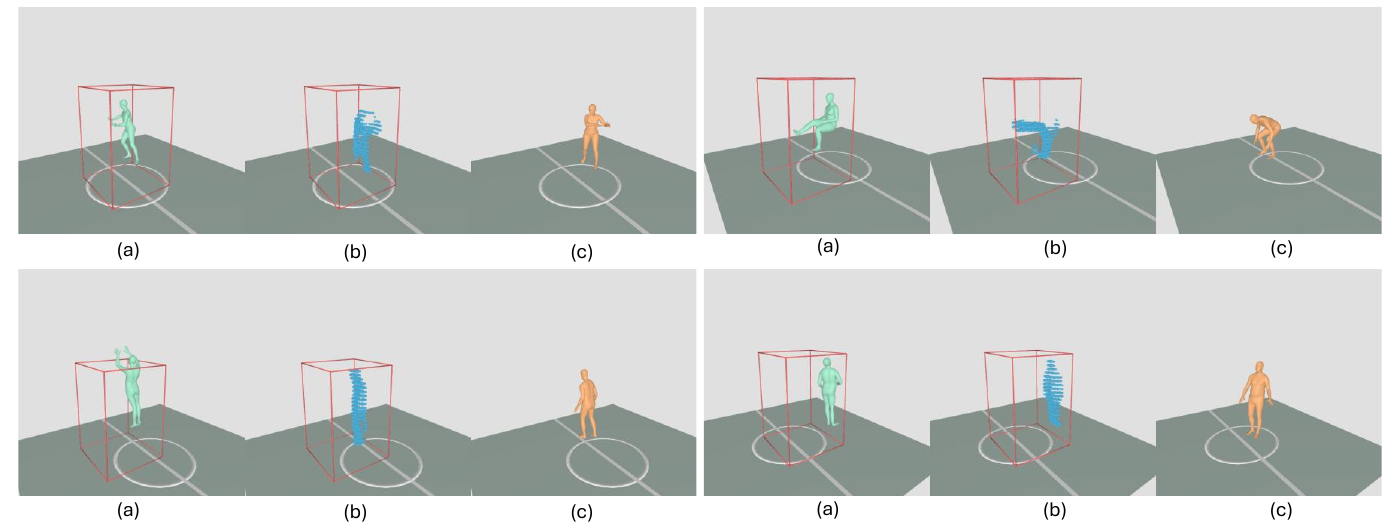}
    \caption{Effect of localization error on the recovered mesh. (a) Ground-truth mesh.
(b) Points extracted by HRE. (c) Recovered mesh. The red box indicates the RoI predicted
by the coarse localization.}
    \label{fig:vis_loc_err}
\end{figure*}

We first analyze how localization error affects the final mesh error.
Figure~\ref{fig:loc_err} shows MVE as a function of localization error, with samples
grouped into bins of increasing error. Up to 250mm, MVE remains flat at
119--122\,mm, a spread comparable to the standard error of the mean. This is a direct
consequence of choosing a sufficiently large crop. Beyond this point, MVE increases
monotonically, reaching 168.7\,mm in the largest bin. Figure~\ref{fig:vis_loc_err} illustrates this failure mode. When the RoI overlaps the body
only partially, the extracted points cover a subset of the body surface and no longer represent the full pose, leading to incorrect reconstruction.

\begin{figure*}
    \centering
    \includegraphics[width=1\linewidth]{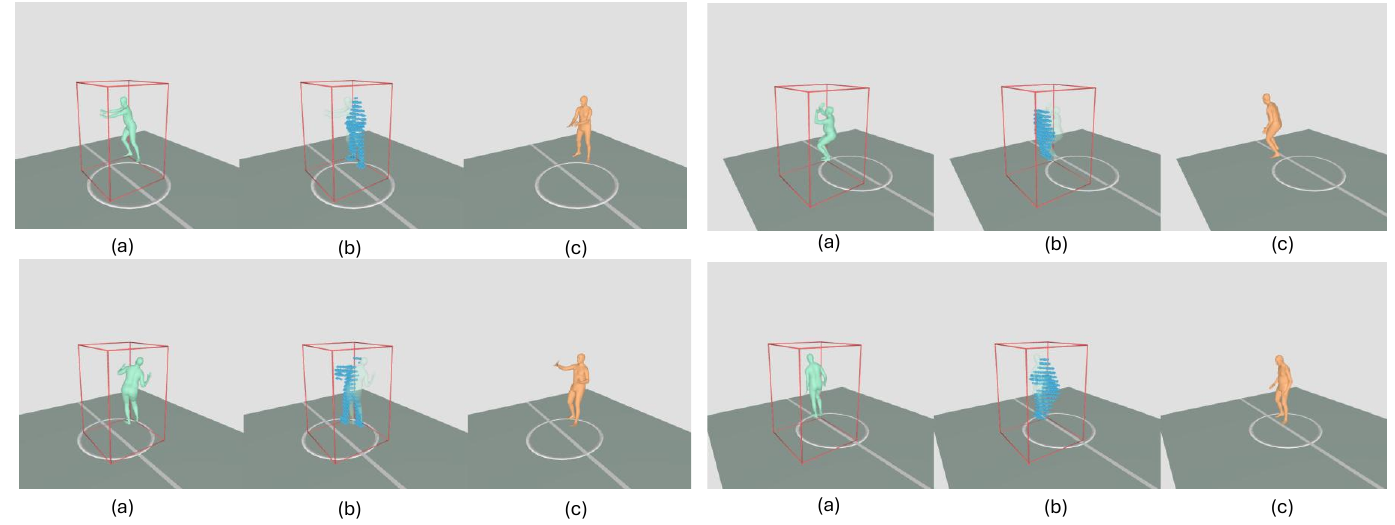}
    \caption{Effect of segmentation error on the recovered mesh. (a) Ground-truth mesh.
    (b) Points extracted by HRE. (c) Recovered mesh.}
    \label{fig:vis_seg_err}
\end{figure*}

When coarse localization is reliable, the next source of failure is segmentation error. We therefore repeat the analysis by grouping well-localized frames into bins of increasing Dice score, shown in Figure~\ref{fig:dice_mve}. MVE decreases as segmentation improves, falling from 336.2\,mm in the lowest bin to 82.2\,mm in the highest, a reduction of 254\,mm. Figure~\ref{fig:vis_seg_err} illustrates this failure mode. The second stage reconstructs
the mesh that best matches the extracted points, so when segmentation is poor the point
set no longer follows the body surface and the recovered mesh is fitted to an incorrect
observation.

\begin{figure*}
    \centering
    \includegraphics[width=1\linewidth]{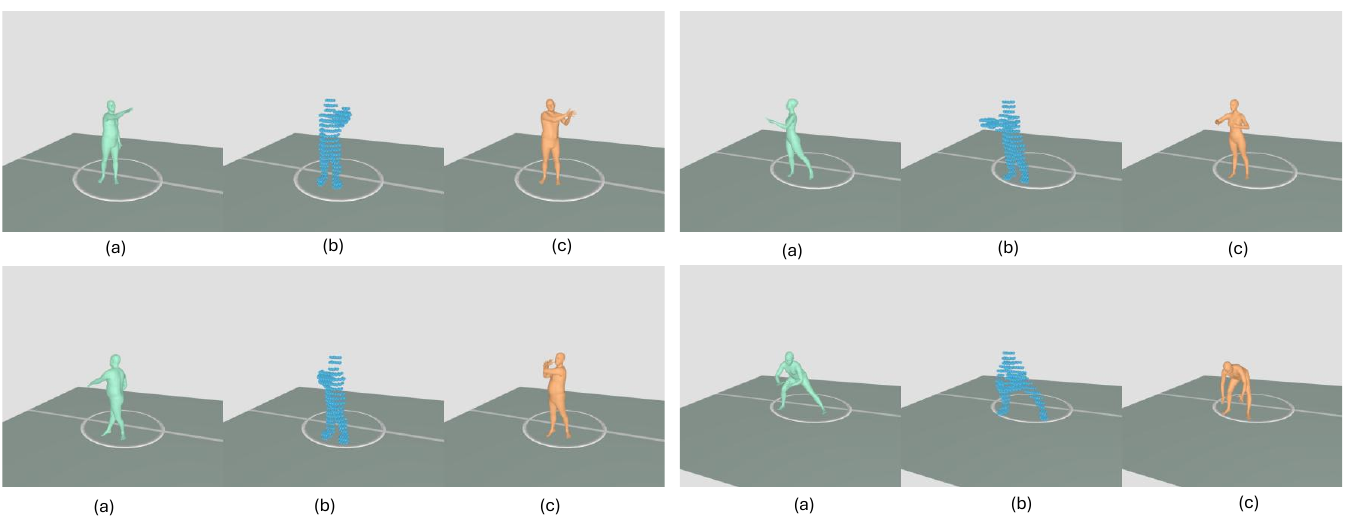}
    \caption{Failure cases of Stage~2. (a) Ground-truth mesh. (b) Points extracted by HRE.
    (c) Recovered mesh.}
    \label{fig:vis_smmr_err}
\end{figure*}

The final source of error arises in Stage~2 (SMMR). As shown in
Figure~\ref{fig:vis_smmr_err}, the recovered mesh can differ from the ground truth even
when the extracted point cloud is accurate, because the same observation is consistent with
more than one body configuration. Extracted point clouds only reflect body surfaces facing the sensor,
so it constrains the visible side of the body while leaving the occluded side
underdetermined. Several distinct poses therefore reproduce the observed points equally
well, and the network may select one that fits the observation but differs from the true configuration.

Among all samples with MVE above 200\,mm, 45.9\% have high localization error
($\ge$300\,mm) and 45.3\% are well localized but have low segmentation quality
(Dice $\le$0.7), leaving 8.8\% attributable to Stage~2. Stage~1 therefore accounts for
91.2\% of severe failures on M4Human, with the two failure modes contributing almost
equally. 
The cross-dataset results show a different pattern. Despite a lower localization error (Table~\ref{tab:hre_milliflow}), MJE
is substantially higher, indicating that most of the error originates in the segmentation
step rather than in coarse localization. This is likely due to the environment shift.
Coarse localization is a global task that depends on the overall spatial distribution of
reflections, which transfers across environments. Segmentation is local, classifying each voxel as body or clutter, and is therefore sensitive to a clutter distribution that changes with the environment. Improving human
reflection extraction, through higher segmentation accuracy and better resilience to
environment shift, is therefore the most promising direction for future research.

\begin{figure}[t]
\centering
\begin{tikzpicture}
\begin{axis}[
    width=\columnwidth,
    height=5.2cm,
    xlabel={Localization error (mm)},
    ylabel={MVE (mm)},
    xmin=-0.6, xmax=8.6,
    ymin=110, ymax=180,
    xtick={0,1,2,3,4,5,6,7,8},
    xticklabels={0--50,50--100,100--150,150--200,200--250,250--300,300--350,350--400,400+},
    xticklabel style={rotate=45, anchor=east, font=\scriptsize},
    ytick={120,140,160,180},
    grid=major,
    grid style={gray!25, dashed},
    tick label style={font=\small},
    label style={font=\small},
    every axis plot/.append style={thick},
]
\addplot+[
    mark=*, mark size=1.4pt,
    error bars/.cd, y dir=both, y explicit,
] coordinates {
    (0,123.0) +- (0,1.43)
    (1,120.7) +- (0,0.55)
    (2,119.8) +- (0,0.37)
    (3,119.7) +- (0,0.34)
    (4,123.3) +- (0,0.37)
    (5,128.5) +- (0,0.46)
    (6,135.0) +- (0,0.61)
    (7,140.6) +- (0,0.76)
    (8,168.7) +- (0,0.90)
};
\end{axis}
\end{tikzpicture}
\caption{Mesh error as a function of localization error. Samples are grouped into 50\,mm
bins.}
\label{fig:loc_err}
\end{figure}

\begin{figure}[t]
\centering
\begin{tikzpicture}
\begin{axis}[
    width=\columnwidth,
    height=5.4cm,
    xlabel={Segmentation Dice},
    ylabel={MVE (mm)},
    xmin=-0.6, xmax=9.6,
    ymin=60, ymax=380,
    xtick={0,1,2,3,4,5,6,7,8,9},
    xticklabels={0.50--0.55,0.55--0.60,0.60--0.65,0.65--0.70,0.70--0.75,0.75--0.80,0.80--0.85,0.85--0.90,0.90--0.95,0.95--1.00},
    xticklabel style={rotate=45, anchor=east, font=\scriptsize},
    ytick={100,150,200,250,300,350},
    grid=major,
    grid style={gray!25, dashed},
    tick label style={font=\small},
    label style={font=\small},
    every axis plot/.append style={thick},
]
\addplot+[
    mark=*, mark size=1.4pt,
    error bars/.cd, y dir=both, y explicit,
] coordinates {
    (0,336.2) +- (0,28.19)
    (1,284.4) +- (0,16.31)
    (2,249.3) +- (0,8.19)
    (3,219.1) +- (0,5.23)
    (4,201.5) +- (0,2.59)
    (5,182.2) +- (0,1.52)
    (6,161.6) +- (0,0.84)
    (7,143.1) +- (0,0.48)
    (8,120.8) +- (0,0.30)
    (9,88.2)  +- (0,0.24)
};
\end{axis}
\end{tikzpicture}
\caption{Mesh error as a function of segmentation quality, restricted to well-localized
frames. Samples are grouped into Dice bins of width 0.05.}
\label{fig:dice_mve}
\end{figure}

\section{Limitations and Discussion on Future Work}
\label{sec:supp_future}

\paragraph{Limitations.} First, the HRE labels are a geometric proxy for radar returns: they mark the radar-visible body surface, whereas actual returns also depend on surface orientation, material, and multipath. Although the proxy agrees well with measured returns (90.1\% of labeled voxels lie within one voxel of a CFAR detection; Sec.~\ref{sec:supp_dataset}), $\mathbf{R}$ should be read as a visible-body likelihood rather than a measured reflection, and the HRE metrics in Tab.~\ref{tab:hre} measure agreement with this proxy. Second, as MilliFlow provides only CFAR point clouds, our cross-dataset evaluation tests joint prediction from pseudo-radar tensors rather than dense mesh recovery from raw tensors.

\paragraph{Future work.} We outline how our framework could be extended to multi-person scenarios and to Doppler-enabled systems. As no public dataset currently provides paired radar tensors and ground-truth meshes for either setting, we leave both as future work.

For multi-person scenarios, a natural extension is to replace the coarse localization head with a person detection module that predicts a variable number of body centers. The segmentation network can then be applied to each detected RoI independently, so Stage 2 remains unchanged and operates on one person at a time.

For Doppler-enabled systems, our pipeline currently operates on the spatial radar tensor and discards the Doppler dimension. Doppler is complementary to the cues we exploit, as it measures radial velocity directly rather than inferring motion from inter-frame differences, and static clutter concentrates near zero Doppler while a moving body does not. It could therefore be used in both stages: as an input channel to the segmentation network, where the velocity signature helps separate body reflections from static background, and as a feature or supervision signal for the motion branch, whose scene-flow objective closely relates to radial velocity. Incorporating this would require a radar tensor retaining the Doppler axis, which the datasets used here do not provide.

\end{document}